\documentclass[10pt,twocolumn,letterpaper]{article}

\usepackage[pagenumbers]{cvpr} % To force page numbers, e.g. for an arXiv version

\usepackage[dvipsnames]{xcolor}

\definecolor{cvprblue}{rgb}{0.21,0.49,0.74}
\usepackage[pagebackref,breaklinks,colorlinks,citecolor=cvprblue]{hyperref}
\usepackage{comment}
\usepackage{tabulary}
\usepackage{multirow}
\usepackage{colortbl}

\newcolumntype{x}[1]{>{\centering\arraybackslash}p{#1pt}}
\newcolumntype{y}[1]{>{\raggedright\arraybackslash}p{#1pt}}
\newcolumntype{z}[1]{>{\raggedleft\arraybackslash}p{#1pt}}
\newlength\savewidth\newcommand\shline{\noalign{\global\savewidth\arrayrulewidth
    \global\arrayrulewidth 1pt}\hline\noalign{\global\arrayrulewidth\savewidth}}
\newcommand{\tablestyle}[2]{\setlength{\tabcolsep}{#1}\renewcommand{\arraystretch}{#2}\centering\footnotesize}

\usepackage{makecell}

\def\method{PointInfinity\xspace}

\def\point{\boldsymbol{p}}
\def\bx{\boldsymbol{x}}
\def\bz{\boldsymbol{z}}

\newcommand{\figref}[1]{Fig.~\ref{fig:#1}}
\newcommand{\tabref}[1]{Table~\ref{tab:#1}}
\newcommand{\secref}[1]{\S\ref{sec:#1}}

\def\paperID{4687} % *** Enter the Paper ID here
\def\confName{CVPR}
\def\confYear{2024}

\title{PointInfinity: Resolution-Invariant Point Diffusion Models}

\author{Zixuan Huang$^{1,2*}$ \quad Justin Johnson$^{1*}$ \quad Shoubhik Debnath$^{1}$ \quad James M. Rehg$^{2}$ \quad Chao-Yuan Wu$^{1*}$\\
$^1$FAIR at Meta, $^2$University of Illinois at Urbana-Champaign
}

\begin{document}

\twocolumn[{%
\renewcommand\twocolumn[1][]{#1}%
\maketitle
\begin{center}
    \centering
    \captionsetup{type=figure}
    \includegraphics[width=1\textwidth]{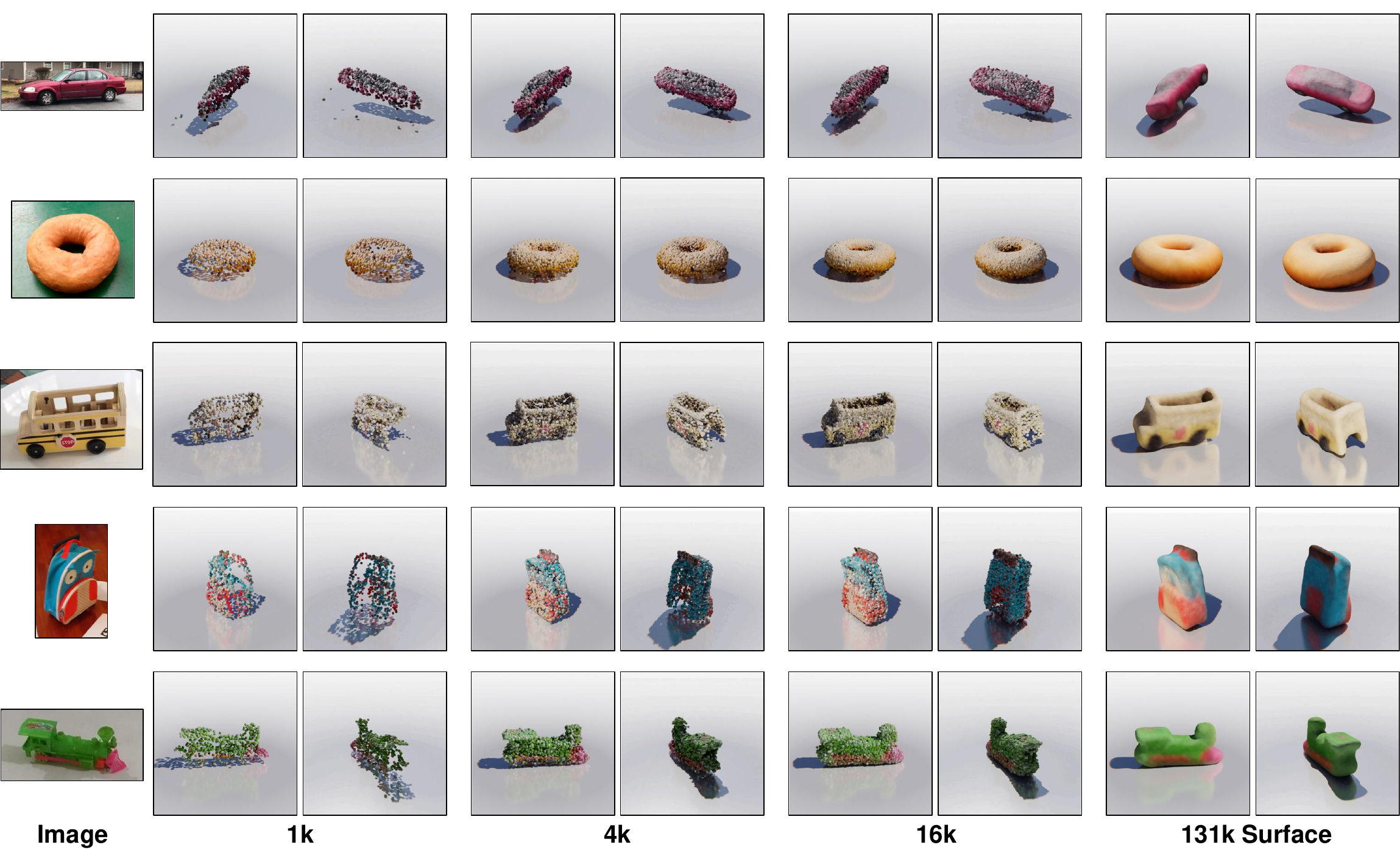}
    % \captionof{figure}{We present a resolution-invariant point diffusion model: our model is trained on \textbf{\emph{low-resolution}} point clouds, but generates
    % \textbf{\emph{high-resolution}} point clouds. Rather than hurting the surface generation quality, our test-time resolution scaling \textbf{\emph{improves}} generated surfaces.}
    \captionof{figure}{We present a resolution-invariant point cloud diffusion model that trains at \textbf{\emph{low-resolution}} (down to 64 points), but generates \textbf{\emph{high-resolution}} point clouds (up to 131k points).
    This test-time resolution scaling \textbf{\emph{improves}} our generation quality.
    We visualize our high-resolution 131k point clouds by converting them to a continuous surface.
    }
\label{fig:teaser}
\end{center}%
}]

\def\thefootnote{*}\footnotetext{Work done at Meta.}\def\thefootnote{\arabic{footnote}}
\begin{abstract}
We present \method, an efficient family of point cloud diffusion models.
Our core idea is to use a transformer-based architecture with a fixed-size, \emph{resolution-invariant} latent representation.
This enables efficient training with low-resolution point clouds, while allowing high-resolution point clouds to be generated during inference.
More importantly, we show that scaling the test-time resolution beyond the training resolution \emph{improves} the fidelity of generated point clouds and surfaces.
We analyze this phenomenon and draw a link to classifier-free guidance commonly used in diffusion models, demonstrating that both allow trading off fidelity and variability during inference.
  Experiments on CO3D show that \method can efficiently generate high-resolution point clouds (up to 131k points, 31$\times$ more than Point-E) with state-of-the-art quality.
\end{abstract}

\section{Introduction}
\label{sec:intro}

Recent years have witnessed remarkable success in diffusion-based 2D image generation~\cite{dhariwal2021diffusion,rombach2022high,saharia2022photorealistic}, 
characterized by unprecedented visual quality and diversity in generated images.
In contrast, diffusion-based 3D point cloud generation methods have lagged behind,
lacking the realism and diversity of their 2D image counterparts.
We argue that a central challenge is the substantial size of typical point clouds: common point cloud datasets~\cite{Geiger2012CVPR,Argoverse2} typically contain point clouds at the resolution of 100K or more.
This leads to prohibitive computational costs for generative modeling due to the quadratic complexity of transformers with respect to the number of input points.
Consequently, state-of-the-art models are severely limited by computational constraints, often restricted to a low resolution of 2048 or 4096 points~\cite{nichol2022point,zhou20213d,luo2021diffusion,zeng2022lion,tyszkiewicz2023gecco}.

In this paper, we propose an efficient point cloud diffusion model that is efficient to train and easily scales to high resolution outputs.
Our main idea is to design a class of architectures with fixed-sized, \emph{resolution-invariant} latent representations.
We show how to efficiently train these models with low resolution supervision, while enabling the generation of high-resolution point clouds during inference.

% intuition
Our intuition comes from the observation that different point clouds of an object can be seen as different samples from a shared continuous 3D surface.
As such, a generative model that is trained to model multiple low-resolution samples from a surface ought to learn a representation from the underlying surface, allowing it to generate high-resolution samples after training.

% introduce the model design
To encode this intuition into model design, 
we propose to decouple the representation of the underlying surface and the representation for point cloud generation.
The former is a constant-sized memory for modeling the underlying surface.
The latter is of variable size, depending on point cloud resolution.
We design lightweight read and write modules for communicating between the two representations.
The bulk of our model's computation is spent on modeling the underlying surface.

% breifly summarize the experiment results
Our experiments demonstrate a high level of resolution invariance with our model\footnote{The resolution-invariance discussed in this paper refers to the property we observe empirically as in experiments, instead of a strict mathematical invariance}.
Trained at a low resolution of 1,024, the model can generate up to 131k points during inference with state-of-the-art quality, as shown in \figref{teaser}.
Interestingly, we observe that using a higher resolution than training in fact leads to slightly \textbf{higher} surface fidelity. We analyze this intriguing phenomenon and draw connection to classifier-free guidance.
We emphasize that our generation output is $>$30$\times$ higher resolution than those from Point-E~\cite{nichol2022point}.
We hope that this is a meaningful step towards scalable generation of \emph{high-quality} 3D outputs.
\section{Related Work}
\label{sec:related}

\paragraph{Single-view 3D reconstruction}
aims to recover the 3D shape given an input image depicting an object or a scene. 
Recent works can be categorized based on the 3D representation they choose.
Commonly used representation includes point clouds~\cite{fan2017point}, voxels~\cite{girdhar2016learning,choy20163d,xie2019pix2vox}, meshes~\cite{groueix2018,wang2018pixel2mesh} and implicit representations~\cite{mescheder2019occupancy,xu2019disn}. 
Results of these works are usually demonstrated on synthetic datasets and/or small-scale real-world datasets such as Pix3D~\cite{pix3d}.
More recently, MCC~\cite{wu2023multiview} proposes to predict occupancy using a transformer-based model.
It shows great zero-shot generalization performance, but it fails to model fine surface details due to its distance-based thresholding~\cite{wu2023multiview}.
Our formulation avoids this issue and generates more accurate point clouds.
Also note that most prior works are regression-based, which leads to deterministic reconstruction, ignoring the multi-modal nature of the reconstruction problem.
Our diffusion-based method generates diverse outputs.

\paragraph{Generative 3D modeling}
learns the distribution of 3D assets, instead of a deterministic mapping. 
Early approaches in this direction often consider modeling 3D generation with GAN~\cite{wu2016learning,achlioptas2018learning,li2018point,shu20193d,valsesia2018learning,hui2020progressive,cai2020learning,gao2022get3d}, normalizing flow~\cite{yang2019pointflow,klokov2020discrete,kim2020softflow} or VAE~\cite{wu2019sagnet,mittal2022autosdf,gao2021tm}. 
More recently, with the success of 2D diffusion models~\cite{dhariwal2021diffusion,rombach2022high}, diffusion-based 3D generative models~\cite{shue20233d,chou2023diffusion,shim2023diffusion,zheng2023locally,hui2022neural,cheng2023sdfusion,li2023diffusion,nam20223d,liu2023meshdiffusion} have been proposed and achieve promising generation quality. Among 3D diffusion models, point cloud diffusion models~\cite{zhou20213d,luo2021diffusion,zeng2022lion,tyszkiewicz2023gecco,nichol2022point} are the most relevant ones to our work. We share the same diffusion framework with these approaches, but propose a novel resolution-invariant method that is both accurate and efficient.
We also goes beyond noise-free synthetic datasets and demonstrate success on more challenging real-world datasets such as CO3D~\cite{reizenstein2021common}.

\paragraph{Transformers}
are widely used in various domains in computer vision~\cite{dosovitskiy2020image,liu2021swin}.
We extend transformers to use a fixed-sized latent representation for a resolution-invariant modeling of 3D point clouds.
The resulting family of architectures includes architectures used in some prior works in recognition and 2D generation~\cite{jaegle2021perceiver,jaegle2021perceiverio,jabri2022scalable},
that were originally designed for joint modeling of multiple modalities.
\section{Background}
\label{sec:background}

\paragraph{Problem Definition.}
The problem studied in this work is RGB-D conditioned point cloud generation, similar to MCC~\cite{wu2023multiview}. Formally, we denote RGB-D images as $I \in \mathbb{R}^{4 \times h \times w}$ and point clouds as $\point \in \mathbb{R}^{n \times 6}$, with 3 channels for RGB and 3 for XYZ coordinates. 
The point clouds we consider in this work can come from various data sources, including the noisy ones from multi-view reconstruction algorithms~\cite{reizenstein2021common}. 

\begin{figure*}[ht]
    \subfloat[\textbf{PointInfinity Overview}\label{fig:method}]{%
    \centering
        \includegraphics[width=0.55\linewidth]{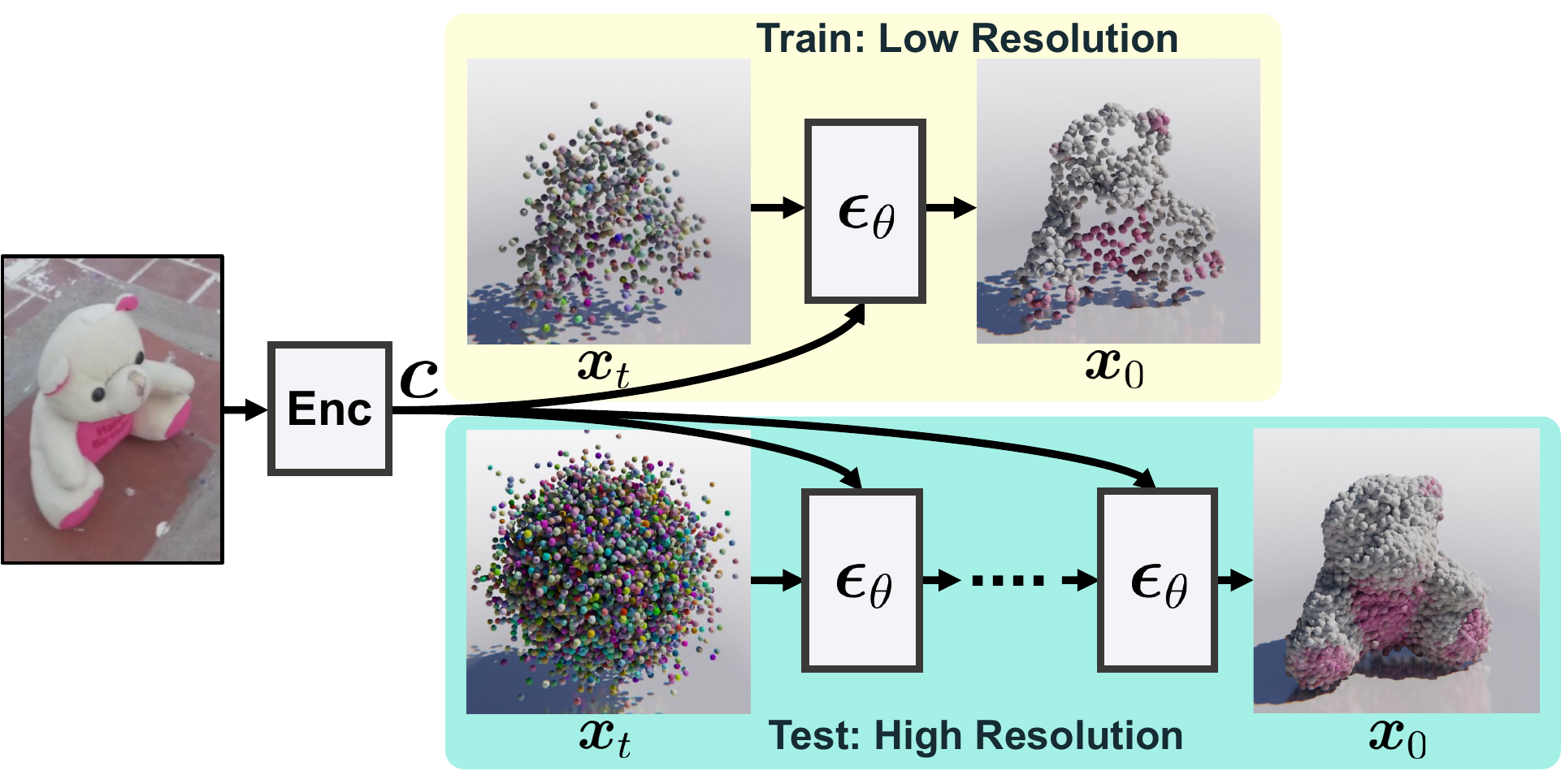}
    }
    \subfloat[\textbf{Denoiser Architecture}\label{fig:block}]{%
    \centering
        \includegraphics[width=0.45\linewidth]{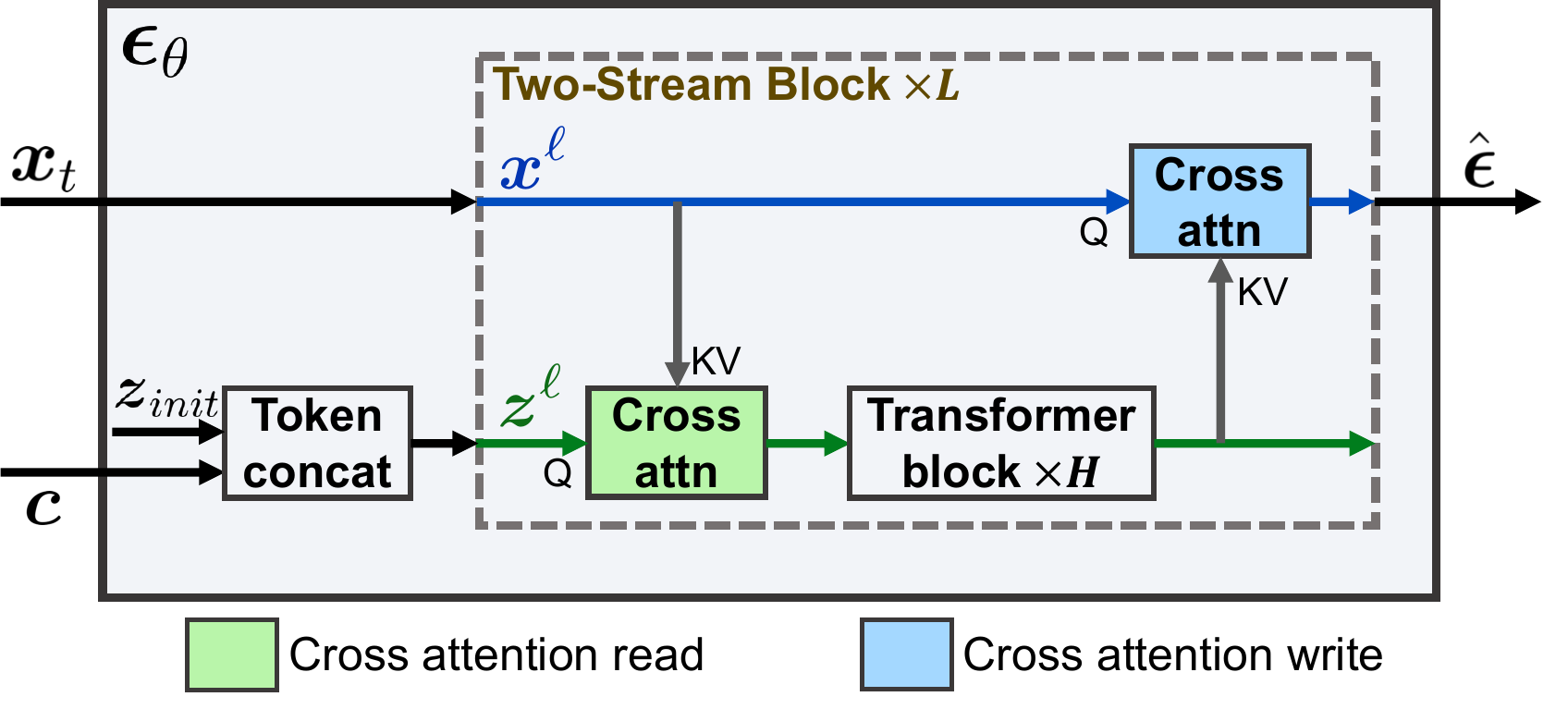}
    }
    \caption{\textbf{Conditional 3D Point Cloud Generation with \method.} (a): At the core of \method is a resolution-invariant conditional denoising model $\boldsymbol{\epsilon}_\theta$. It uses low-resolution point clouds for training and generates high-resolution point clouds at test time.
    (b): The main idea is a ``Two-Stream" transformer design that decouples a fixed-sized latent representation $\bz$ for capturing the underlying 3D shape and a variable-sized data representation $\bx$ for modeling of the point could space.
        `Read' and `write' cross-attention modules are used to communicate between the two streams of processing.
        Note that most of the computation happens in the \emph{latent stream} for modeling the underlying shape. This makes it less susceptible to the effects of point cloud resolution variations.}
    \label{fig:method}

\end{figure*}

\paragraph{Denoising Diffusion Probabilistic Model (DDPM).}
Our method is based on the DDPM~\cite{ho2020denoising}, which consists of two processes: 1) the diffusion process which destroys data pattern by adding noise, and 2) the denoising process where the model learns to denoise. At timestep $t \in [0, T]$, the diffusion process blends Gaussian noise $\boldsymbol{\epsilon} \sim \mathcal{N}(\mathbf{0},\mathbf{I})$ with data sample $\point_0$ as % \sim p(\boldsymbol{x}_0)$
\begin{equation}
    \point_t = \sqrt{\bar{\alpha_t}} \point_0 + \sqrt{1 - \bar{\alpha_t}} \boldsymbol{\epsilon},
\end{equation}
where $\bar{\alpha_t}$ denotes the noise schedule. The denoiser $\boldsymbol{\epsilon}_\theta (\point_t, t)$ then learns to recover the noise from $\point_t$ with loss
\begin{equation}
    L_{simple}(\theta) = \mathbb{E}_{t,\point_0,\boldsymbol{\epsilon}} \lVert \boldsymbol{\epsilon} - \boldsymbol{\epsilon}_\theta (\point_t, t) \rVert_2^2.
\end{equation}
During inference, we use the stochastic sampler proposed in Karras et al.~\cite{karras2022elucidating} to generate samples. 

\paragraph{Classifier-Free Guidance.}
Conditional diffusion models often use classifier-free guidance~\cite{ho2022classifier} to boost the sample quality at the cost of sample diversity. During training, the condition of the model is dropped with some probability and the denoiser will learn to denoise both with and without condition. At test time, we linearly combine the conditional denoiser with unconditional denoiser as follows
\begin{equation}
    \Tilde{\boldsymbol{\epsilon}_\theta} (\point_t, t | \boldsymbol{c}) = (1 + \omega) \boldsymbol{\epsilon}_\theta (\point_t, t | \boldsymbol{c}) - \omega \boldsymbol{\epsilon}_\theta (\point_t, t),
\end{equation}
where $\omega$ is the classifier-free guidance scale and $\Tilde{\boldsymbol{\epsilon}_\theta} (\point_t, t | \boldsymbol{c})$ is the new denoiser output.

\paragraph{Transformer-based~\cite{vaswani2017attention} point diffusion models}
have been widely used in prior works~\cite{nichol2022point}, due to its permutation equivariant nature.
Namely, when we permute the input noisy point cloud, transformers guarantee that the output noise predictions are also permuted in the same way. 

However, as we will show in \secref{exp}, vanilla transformers are not resolution-invariant ---
Testing with a different resolution from training significantly reduces accuracy.
Furthermore, they scale quadratically w.r.t.\ to resolution, making them unamenable for high-resolution settings.
To generate denser outputs, Point-E~\cite{nichol2022point} trains a separate upsampler for upsampling points from 1024 to 4096.
In the next section, we will show how to scale the resolution to up to 131k points without a separate upsampler.
\section{Point Cloud Generation with \method}
\label{sec:method}
The main idea of \method is a resolution-invariant model, with which we train the model efficiently using low-resolution point clouds, while still supporting point cloud generation at a higher resolution.
\figref{method} illustrates an overview of the system.

\subsection{Model}
To achieve resolution invariance, we propose to parameterize
$\epsilon_\theta(\point_t, t | c)$
to be a \emph{2-stream} transformer-based model.
The model first linearly projects noisy input points $\point_t$ into representations $\bx_t$.
Then a stack of $L$ two-stream blocks process $\bx_t$ and finally predicts $\hat{\boldsymbol{\epsilon}}$.

\paragraph{The Two-Stream Block.}
The main idea of our two-stream block is to introduce a fixed-sized latent representation $\bz$ for capturing the underlying 3D shape and a \emph{latent} processing stream for modeling it.
Concretely, the $\ell$-th block takes in two inputs $\bx^\ell \in \mathbb{R}^{n\times d}$, $\bz^\ell \in \mathbb{R}^{m\times d}$ and outputs $\bx^{(\ell+1)} \in \mathbb{R}^{n\times d}$, $\bz^{(\ell+1)} \in \mathbb{R}^{m\times d}$.
At the first two-stream block ($\ell$ = 0), the data-stream $\bx^0$ is fed with the noisy point cloud $\bx_t$. 
% The data input of the first block $\bx^0$ is $\bx_t$. 
The latent input of the first block $\bz^0$ is a learned embedding $\boldsymbol{z}_\mathrm{init}$ cancatenated with conditioning tokens $c$ in the token dimension.

Within each two-stream block, we will first use a \emph{read} cross attention block to cross attend information from data representation $\bx^\ell$ into the latent representation $\bz^\ell$,
\begin{align}
\tilde{\bz}^\ell := \mathrm{CrossAttn}(\bz^\ell, \bx^\ell, \bx^\ell),
\end{align}
where $\mathrm{CrossAttn}(Q, K, V)$ denotes a cross attention block with query $Q$, key $K$, and value $V$.
Then we use $H$ layers of transformer blocks to model the latent representation
\begin{align}
\bz^{(\ell+1)} := \mathrm{Transformer}(\tilde{\bz}^\ell)
\end{align}
Finally, we will use a \emph{write} cross attention block to write the latent representation back into the data stream through
\begin{align}
\bx^{(\ell+1)} := \mathrm{CrossAttn}(\bx^\ell, \bz^{(\ell+1)}, \bz^{(\ell+1)})
\end{align}
\figref{block} illustrates our design.
Note that the \emph{latent stream} processes tokens that are fixed-sized,
while the \emph{data stream} processes variable-sized tokens projected from noisy point cloud data.
Since the bulk of the computation is spent on the fixed-sized latent stream, the processing is less affected by the resolution of the data stream.
Also note that with this design, the computation only grows linearly  with the size of $\bx$, instead of growing quadratically.

\subsection{Implementation Details}
\paragraph{Architecture Details.}
We use $L=6$ two-stream blocks in our denoiser, each includes $H=4$ transformer blocks. For conditioning, we use the MCC encoder~\cite{wu2023multiview} to encode the RGB-D image into 197 tokens, and we use the time step embedding in~\cite{nichol2022point} to encode time step $t$ as a vector. Concatenating these two along the token dimension, we obtain the condition tokens $c$ consisting of 198 vectors of dimension $d=256$. $\boldsymbol{z}_\mathrm{init}$ consists of 256 tokens, so the latent representation $\bz^\ell$ has $m=454$ tokens in total. The default training resolution $n_\mathrm{train}$ we use is 1024, while the test-time resolution $n_\mathrm{test}$ we consider in the experiments varies from 1024 to 131,072. 

\paragraph{Training Details.}
We train our model with the Adam~\cite{kingma2014adam} optimizer. We use a learning rate of $1.25\times 10^{-4}$, a batch size of 64 and momentum parameters of (0.9, 0.95). We use a weight decay of 0.01 and train our model for 150k iterations on CO3D. For diffusion parameters, we use a total of 1024 timesteps with the cosine noise scheduler. We also use latent self-conditioning of probability 0.9 during training following~\cite{jabri2022scalable}.

\paragraph{Surface Extraction.}
Because our model is able to generate high-resolution point clouds, it is possible to directly extract surface from the generated point clouds. To do so, we first create a set of 3D grid points in the space. For each point, we find the neighbor points in the point cloud and compute the mean distance to these points. We then use the marching cube~\cite{lorensen1998marching} to extract the surface by thresholding the mean distance field.
\section{Experiments}
\label{sec:exp}

\subsection{Dataset}
\paragraph{CO3D.} 
We use CO3D-v2~\cite{reizenstein2021common} as our main dataset for experiments. 
CO3D-v2 is a large real-world collection of 3D objects in the wild, that consists of $\sim$37k objects from 51 object categories.
The point cloud of each object is produced by COLMAP~\cite{schonberger2016structure,schonberger2016pixelwise} from the original video capture.
Despite the noisy nature of this process, we show that our model produces faithful 3D generation results.

\subsection{Evaluation Protocol}
\label{subsec:protocal}

\paragraph{Metrics.} 
Following~\cite{mescheder2019occupancy,wu2023multiview,huang2023shapeclipper}, the main evaluation metric we use for RGB-D conditioned shape generation is Chamfer Distance (CD). 
Given the predicted point cloud $S_1$ and the groundtruth point cloud $S_2$, CD is defined as an average of accuracy and completeness:
\begin{equation}
    \small
    d(S_1, S_2) = \frac{1}{2|S_1|}\sum_{x \in S_1} \min_{y \in S_2} \|x-y\|_2 + \frac{1}{2|S_2|}\sum_{y \in S_2} \min_{x \in S_1} \|x-y\|_2
    \label{eq:chamfer}
\end{equation}
Another metric we consider is F-score, which measures the alignment between the predicted point cloud and the groundtruth under a classification framing. Intuitively, it can be understood as the percentage of surface that is correctly reconstructed. In our work, we use a threshold of 0.2 for all experiments --- if the distance between a predicted point and a groundtruth point is less than 0.2, we consider it as a correct match.

In addition to shape evaluation metrics, we also consider peak signal-to-noise ratio (PSNR) for texture evaluation.

\paragraph{Protocol.} 
Note that point clouds with more points might be trivially advantageous in \emph{completeness}, and thus Chamfer Distance or F-score.
Consequently, in this paper we compute CD not only on the traditional \emph{full point cloud} setting (denoted `CD@full'), but also the \emph{subsampled} setting (1024 points by default; denoted `CD@1k') to ensure all methods are compared under the same number of points.
Intuitively, `CD@1k' measures the `surface quality' under a certain resolution.\footnote{For F-score, we always report the subsampled version.} 
In addition, all objects are standardized such that they have zero mean and unit scale to ensure a balanced evaluation across all objects.

\subsection{Baselines}
We compare \method with two SOTA models, Multiview Compressive Coding (MCC)~\cite{wu2023multiview} and Point-E~\cite{nichol2022point}. 

\paragraph{MCC~\cite{wu2023multiview}} studies the problem of RGB-D conditioned shape reconstruction and learns implicit reconstruction with regression losses. MCC and our model use the same RGB-D encoder and both use CO3D-v2 as training set. 
One main difference between MCC and our model is that MCC uses a deterministic modeling and does not model interactions between query points. 

\paragraph{Point-E~\cite{nichol2022point}}
is a point cloud diffusion model using a vanilla transformer backbone. 
As the official training code is not released, we report results based on our reimplementation.
We use the same RGB-D encoder as our method for fair comparison. 
The main difference between Point-E and \method lies the architecture of the diffusion denoisers.

\begin{table}[t]
	\footnotesize
    \centering
	\tablestyle{5.0pt}{1.05}
    \begin{tabular}{@{}llx{25}x{25}x{25}x{25}@{}}
    Metric & Method  & 1024 & 2048 & 4096 & 8192 \\ \shline
    \multirow{2}{*}{CD@1k ($\downarrow$)}   
    & Point-E~\cite{nichol2022point} & 0.239 & 0.213 & 0.215 & 0.232 \\ %\cline{2-6} 
    & \textbf{Ours}    & \cellcolor{green!10} \textbf{0.227} & \cellcolor{green!30} \textbf{0.197} & \cellcolor{green!50} \textbf{0.186} & \cellcolor{green!60} \textbf{0.181} \\ \hline
    \multirow{2}{*}{CD@full ($\downarrow$)} 
    & Point-E~\cite{nichol2022point} & 0.239 & 0.200 & 0.194 & 0.205 \\ %\cline{2-6}
    & \textbf{Ours}    & \cellcolor{green!20} \textbf{0.227} & \cellcolor{green!40} \textbf{0.185} & \cellcolor{green!60} \textbf{0.164} & \cellcolor{green!80} \textbf{0.151} \\ \hline
    \multirow{2}{*}{PSNR ($\uparrow$)} 
    & Point-E~\cite{nichol2022point} & 13.31 & 13.46 & 13.28 & 12.60 \\ %\cline{2-6} 
    & \textbf{Ours}    & \cellcolor{green!5} \textbf{13.37} & \cellcolor{green!20} \textbf{13.88} & \cellcolor{green!30} \textbf{14.15} & \cellcolor{green!40} \textbf{14.27}
    \end{tabular}
    \caption{\textbf{Effect of Test-Time Resolution Scaling.} Here we compare \method and Point-E~\cite{nichol2022point} at different testing resolutions $n_\textrm{test}$. With \method, using a higher resolution during testing does not only lead to denser capture of the surface, it also improves the surface quality, as reflected by CD@1k and PSNR. 
    On the contrary, Point-E, which uses a vanilla transformer backbone, sees a performance drop at high resolution.}
    \label{tab:scaling-perf}
\vspace{8mm}

\footnotesize
\centering
\begin{tabular}{@{}lx{35}x{35}x{35}x{35}@{}}
Resolution & 1024 & 2048 & 4096 & 8192 \\ \shline
CD@1k ($\downarrow$) & 0.405 & \cellcolor{green!10} 0.372 & \cellcolor{green!30}0.352 & \cellcolor{green!50}\textbf{0.343} \\ 
FS ($\uparrow$) & 0.336 & \cellcolor{green!15}0.376& \cellcolor{green!30}0.398 & \cellcolor{green!50}\textbf{0.409} \\ 
PSNR ($\uparrow$) & 10.94 & \cellcolor{green!5}11.39 & \cellcolor{green!20} 11.63 & \cellcolor{green!30}\textbf{11.75} \\
\end{tabular}
\caption{\textbf{Generalization to the RGB condition.} Here we evaluate \method trained only with RGB condition at different testing resolutions $n_\textrm{test}$. We observe a similar performance improving trend with higher test-time resolutions.}
\label{tab:rgb-scaling-perf}
\vspace{8mm}

\footnotesize
\centering
\begin{tabular}{@{}lx{35}x{35}x{35}x{35}@{}}
Resolution & 1024 & 2048 & 4096 & 8192 \\ \shline
CD@1k ($\downarrow$) & 0.251 & \cellcolor{green!10} 0.213 & \cellcolor{green!30} 0.203 & \cellcolor{green!50} \textbf{0.197} \\ 
CD@full ($\downarrow$) & 0.251 & \cellcolor{green!20}0.199 & \cellcolor{green!40}0.177 & \cellcolor{green!60}\textbf{0.163} \\ 
PSNR ($\uparrow$) & 13.09 & \cellcolor{green!10} 13.63 & \cellcolor{green!25} 13.85 & \cellcolor{green!40} \textbf{13.97} \\
\end{tabular}
\caption{\textbf{Generalization to Different Backbone Variants.} Our two-stream transformer design include a wide range of variants, including the PerceiverIO~\cite{jaegle2021perceiverio} architecture originally designed for fusing different input modalities for recognition. We observe a similar performance-improving property of test-time resolution scaling with this backbone variant as well. }
\label{tab:perceiver-scaling-perf}
\end{table}

\subsection{Main Results}
\label{sec:main_results}

\paragraph{Test-Time Resolution Scaling.} 
\tabref{scaling-perf} compares performance of \method at different testing resolutions $n_\mathrm{test}$.
As we can see, despite that the $n_\mathrm{test} \neq n_\mathrm{train}$, increasing test-time resolution in fact slightly \emph{improves} the generated surface quality, as reflected on CD@1k.
This verifies the resolution invariance property of \method.
We hypothesize the slight improvement comes from that the read operator gets to incorporate more information into the latent representation, leading to better modeling of the underlying surface.
In \secref{analysis}, we will provide a more detailed analysis.
On the contrary, the performance of Point-E~\cite{nichol2022point} \emph{decreases} with higher testing resolution.
This is expected, as unlike \method, the size of Point-E~\cite{nichol2022point}'s latent representations changes with the resolution, affecting the behavior of all attention operations, making it \emph{not} resolution-invariant.

\paragraph{Generalization Analysis.}
Here we analyze how \method generalizes to different settings like different conditions and backbones.
\tabref{rgb-scaling-perf} presents results on a different condition. Specifically, we explore whether our finding generalizes to the “RGB-conditioned” point generation task.
We can see that when only conditioned on RGB images, \method similarly demonstrates strong resolution invariance.
Performance evaluated on all three metrics improves as test-time resolution $n_\mathrm{test}$ increases.

Note that our default implementation based on~\cite{jabri2022scalable} represents only one instance of the two-stream family.
The PerceiverIO~\cite{jaegle2021perceiverio} architecture originally designed for fusing different input modalities for recognition is another special case of a two-stream transformer model.
The main difference between our default architecture and PerceiverIO lies in the number of read-write cross attention.
\tabref{perceiver-scaling-perf} presents scaling behaviors with PerceiverIO.
We can see that as expected, the performance similarly improves as the test-time resolution increases. 
This verifies that our findings generalize to other backbones within the two-stream family.

\paragraph{SOTA Comparisons.}
We then compare \method with other state-of-the-art methods on CO3D, including MCC~\cite{wu2023multiview} and Point-E~\cite{nichol2022point}. We report the result under a test-time resolution of 16k for our method. As shown in~\tabref{sota}, our model outperforms other SOTA methods significantly.
\method achieves not only better surface generation fidelity (9\% better than Point-E and 24\% better than MCC quantified by CD@1k), but also generates better texture (as shown in better PSNR). 
\begin{table}
    \tablestyle{1.0pt}{1.05}
    \begin{tabular}{@{}lx{50}x{50}x{50}@{}}
    Method & CD@1k ($\downarrow$) & FS ($\uparrow$) & PSNR ($\uparrow$) \\
    \shline
    MCC~\cite{wu2023multiview} & 0.234 &  0.549 & 14.03 \\
    Point-E~\cite{nichol2022point} & 0.197 & 0.675 & 14.25 \\
    \textbf{\method}\quad\quad\quad & \textbf{0.179} & \textbf{0.724} & \textbf{14.31} \\
    \end{tabular}
    \caption{\textbf{Comparison with Prior Works.} We see that \method outperforms other state-of-the-art methods significantly on all metrics we evalute, 
    demonstrating the effectiveness our resolution-invariant point diffusion design.}
    \label{tab:sota}
\end{table}

\paragraph{Comparisons with Unconditional Models.}
Additionally, we compare \method with unconditional 3D generative models in terms of resolution-invariance. Specifically, we consider Point-Voxel Diffusion (PVD)~\cite{luo2021diffusion} and Gradient Field (ShapeGF)~\cite{cai2020learning}. These models are originally designed for unconditional 3D shape generation (no color), and are trained with different resolutions and data. Therefore, we report relative metrics when comparing with them, so that numbers between different methods are comparable. The results of relative CD are shown in~\cref{tab:uncond-compare}. We observe that as resolution increases, PointInfinity's performance improves, while ShapeGF's performance remains almost unchanged. On the other hand, PVD's performance significantly drops. 
This verifies the superior resolution-invariance property of \method, even when compared to models designed for different 3D generation scenarios.

\begin{table}[h]
    \footnotesize
    \centering
    \begin{tabular}{@{}lx{35}x{35}x{35}x{35}@{}}
    Resolution & 1$\times$ & 2$\times$ & 4$\times$ & 8$\times$ \\ \shline
    PVD~\cite{luo2021diffusion} & 1.000 & 3.605 & 4.290 & 4.221 \\ 
    GF~\cite{cai2020learning} & 1.000 & 0.999 & 1.000 & 0.999 \\ 
    \textbf{PointInfinity} & 1.000 & \textbf{0.868} & \textbf{0.819} & \textbf{0.797} \\
    \end{tabular}
    \caption{\textbf{Comparison with Unconditional Models.} We see that \method outperforms other unconditional 3D generative methods, including PVD and ShapeGF, in terms of resolution-invariance.}
    \label{tab:uncond-compare}
\end{table}

\subsection{Complexity Analysis} 
\begin{figure*}[t]
\subfloat[\textbf{Train Iter Time}\label{fig:co3d-scaling-comp-train}]{%
\centering
	\includegraphics[width=0.25\linewidth]{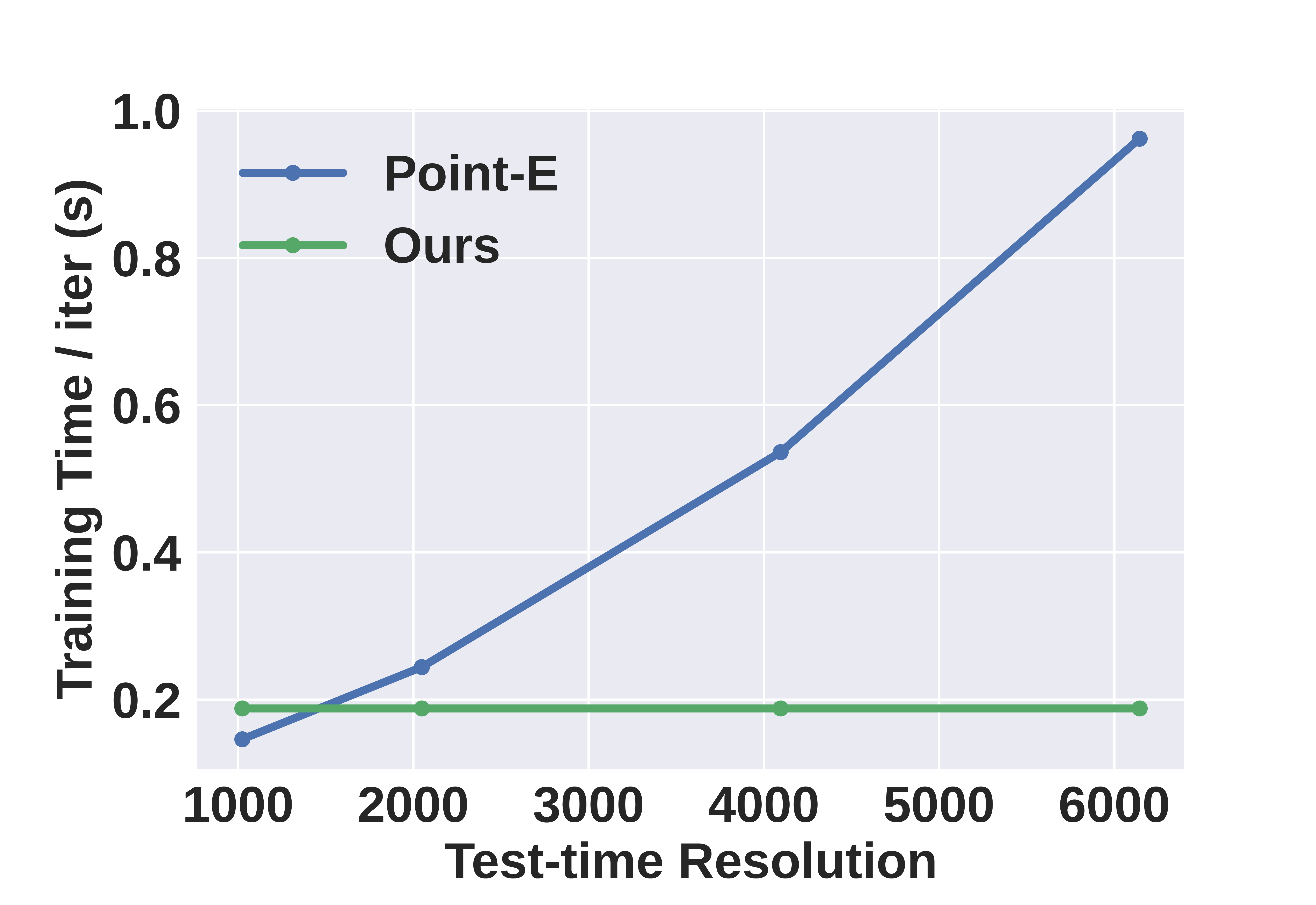}
}
\subfloat[\textbf{Train Memory}\label{fig:co3d-scaling-mem-train}]{%
\centering
	\includegraphics[width=0.25\linewidth]{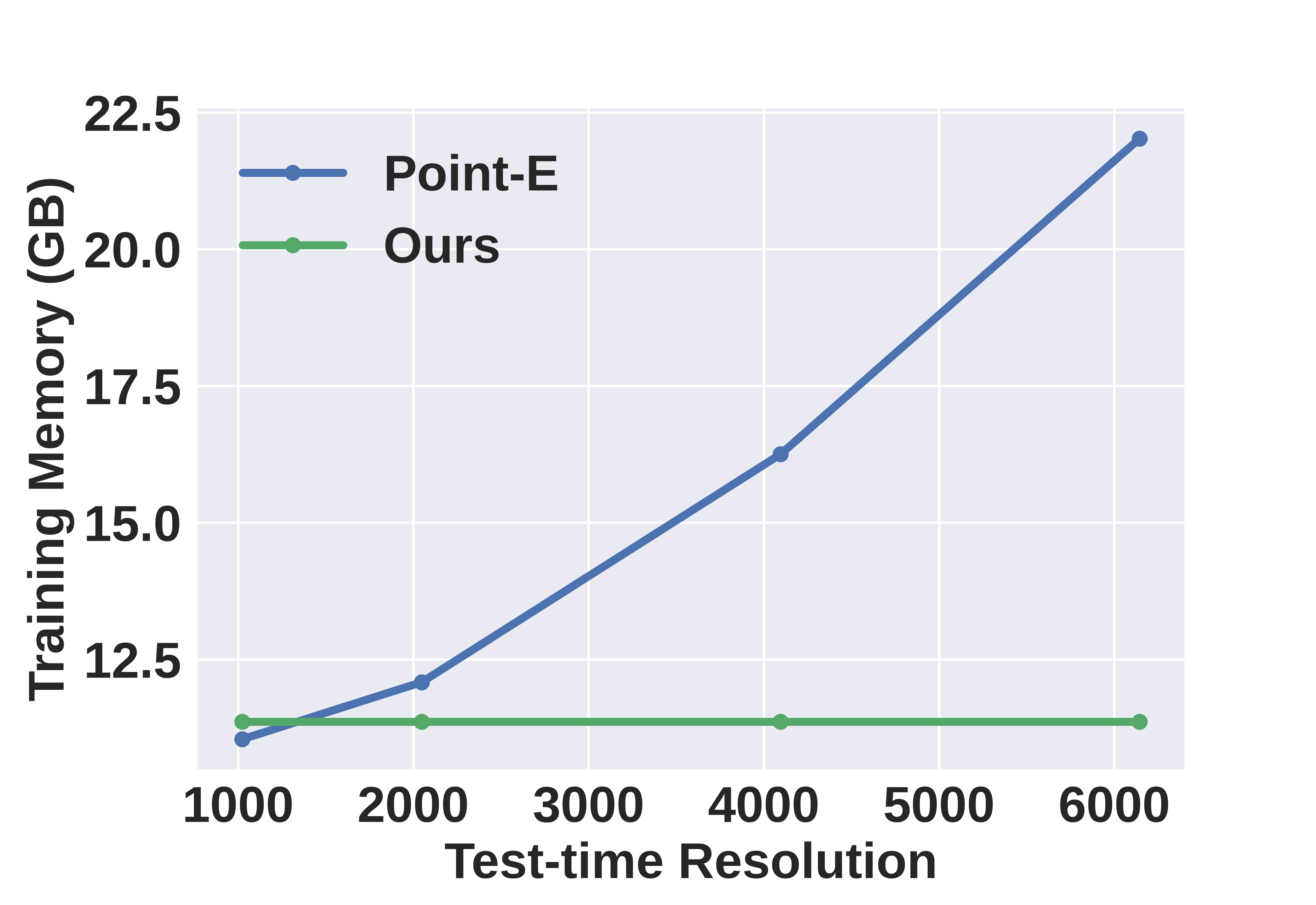}
}
\subfloat[\textbf{Inference Iter Time}\label{fig:co3d-scaling-comp-test}]{%
\centering
	\includegraphics[width=0.25\linewidth]{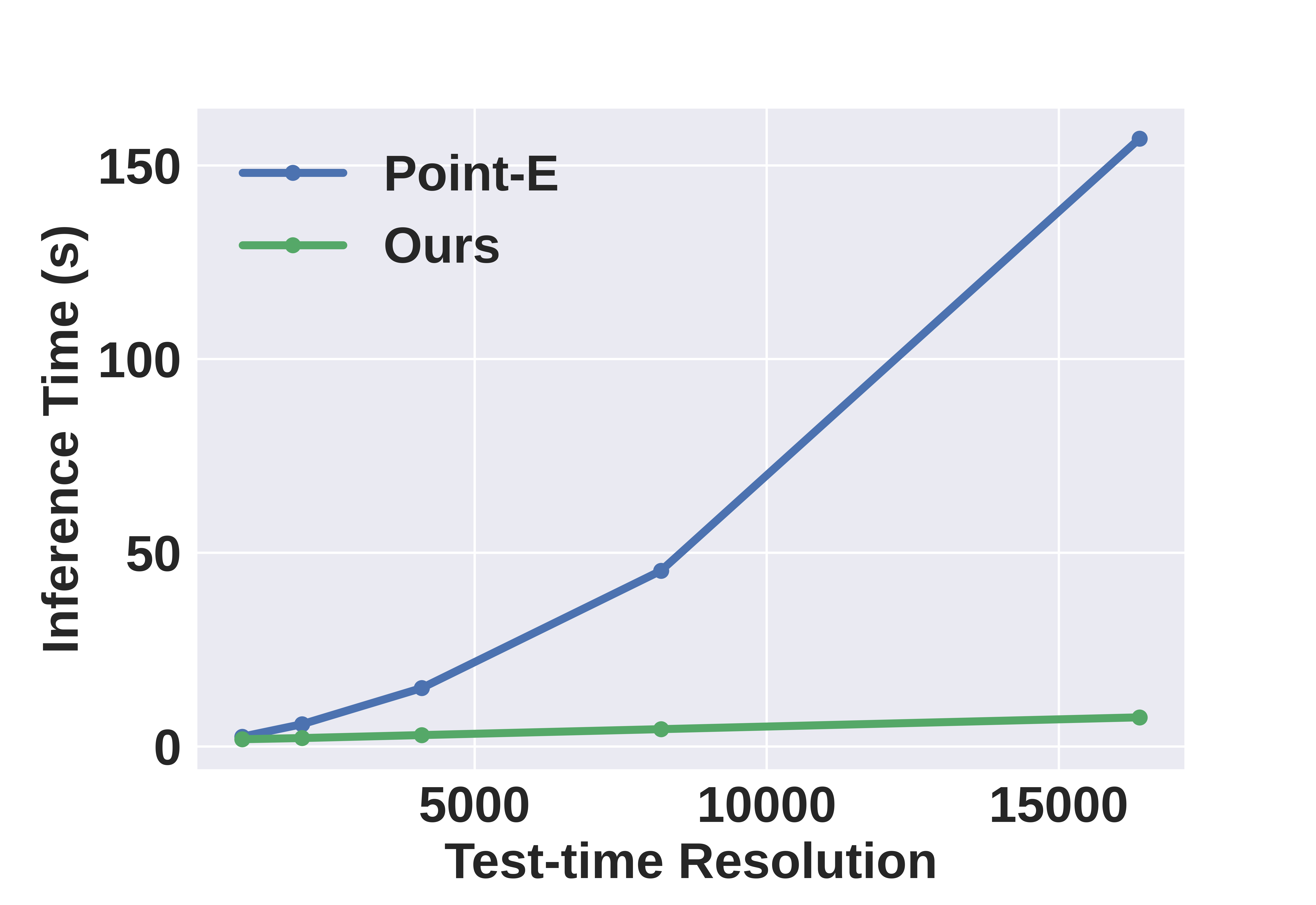}
}
\subfloat[\textbf{Inference Memory}\label{fig:co3d-scaling-mem-test}]{%
\centering
	\includegraphics[width=0.25\linewidth]{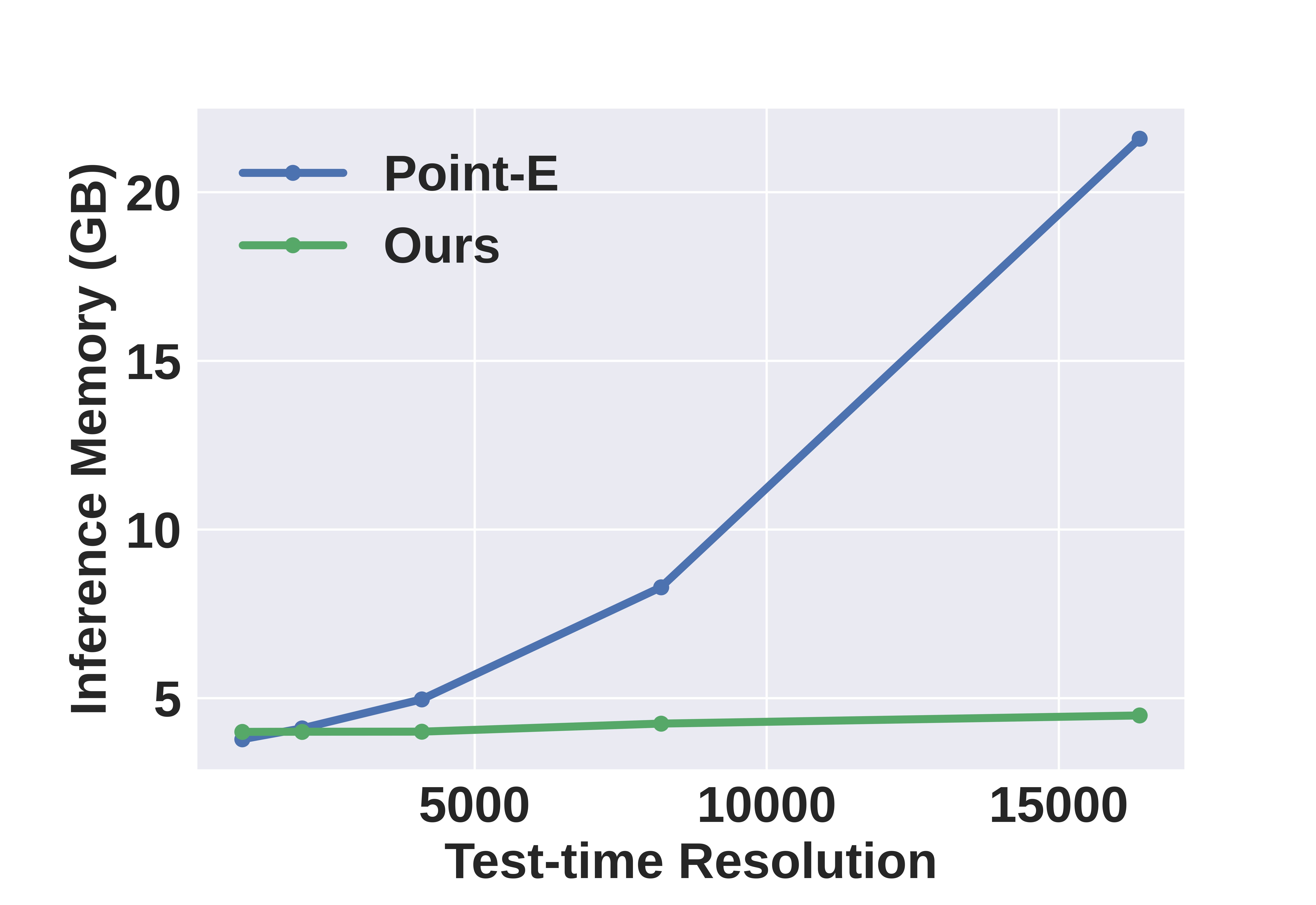}
}
\caption{\textbf{\method scales favorably compared to Point-E~\cite{nichol2022point} in both computation time and memory for both training and inference.} (a,b): Thanks to the resolution-invariant property of \method, the training iteration time and memory stays constant regardless of the test-time resolution $n_\mathrm{test}$. Point-E on the other hand requires $n_\mathrm{train} = n_\mathrm{test}$ and scales quadratically. (c,d): Our inference time and memory scales linearly with respect to $n_\mathrm{test}$ with our two-stream transformer design, while Point-E scales quadratically with the vanilla transformer design.\vspace{2mm}}\label{fig:complexity}
\end{figure*}

\begin{table*}[t]
\subfloat[\textbf{Training Resolution}\label{tab:ablation:train-reso}]{%
\tablestyle{1.0pt}{1.05}
\begin{tabular}{@{}lx{33}x{33}x{33}@{}}
$n_\mathrm{train}$ & CD@1k($\downarrow$) & FS($\uparrow$) & PSNR($\uparrow$) \\
\shline
64 & \underline{0.178} & 0.722 & 14.28 \\
256 & \underline{0.174} & \underline{0.737} & \underline{14.41} \\
1024 (default) & \underline{0.179} & \underline{0.724} & \underline{14.31} \\
2048 & 0.183 & 0.708 & 14.19 \\
\\
\end{tabular}
}\hfill%
\subfloat[\textbf{Number of Latent Tokens}\label{tab:ablation:latent}]{%
\tablestyle{1.0pt}{1.05}
\begin{tabular}{@{}lx{33}x{33}x{33}@{}}
$z_\mathrm{init}$ dim & CD@1k($\downarrow$) & FS($\uparrow$) & PSNR($\uparrow$) \\
\shline
64 & 0.457 & 0.262 & 10.90 \\
128 & 0.182 & 0.719 & 14.25 \\
256 (default) & \underline{0.179} & \underline{0.724} & \underline{14.31} \\
512 & \underline{0.176} & \underline{0.729} & \underline{14.45} \\
\\
\end{tabular}
}\hfill
\subfloat[\textbf{Mixture Baseline}\label{tab:ablation:mixture}]{%
\tablestyle{1.0pt}{1.05}
\begin{tabular}{@{}lx{33}x{33}x{33}x{33}@{}}
&$n_\mathrm{test}$ & CD@1k($\downarrow$) & FS($\uparrow$) & PSNR($\uparrow$) \\
\shline
Mixture &1024 & 0.227 & 0.622 & 13.37 \\
Mixture &2048 & 0.220 & 0.619 & 13.21 \\
Mixture &4096 & 0.215 & 0.625 & 13.12 \\
Mixture &8192 & 0.211 & 0.632 & 13.07 \\
\textbf{\method} &8192& \textbf{0.181} & \textbf{0.721} & \textbf{14.27}
\end{tabular}
}\hfill
\caption{\textbf{Ablation Experiments on CO3D-v2.} We perform ablations on the CO3D-v2 dataset~\cite{reizenstein2021common}. Specifically, we study the impact of training resolution (a), the size of the latent representations (b), and verify the advantage of \method over a `mixture' baseline for generating high resolution point clouds.}
\label{table:ablation}
\end{table*}

We next analyze the computational complexity of \method at different test-time resolutions.
The computational analysis in this section is performed on a single NVIDIA GeForce RTX 4090 GPU with a batch size of 1. 
Thanks to the resolution-invariance property, \method can generate point clouds of different test-time resolutions $n_\mathrm{test}$ without training multiple models.
On the other hand, Point-E~\cite{nichol2022point} requires the training resolution to match with the testing resolution, since it is resolution specific.
We present detailed benchmark results comparing the iteration time and memory for both training and testing in~\figref{complexity}.
We can see that the training time and memory of Point-E model scales \emph{quadratically} with test-time resolution, while our model remains \emph{constant}. 
Similarly at test time, Point-E scales quadratically with input resolution, while our inference computation scales \emph{linearly}, thanks to our two-stream design.

\begin{figure}[h]
\includegraphics[width=0.495\linewidth]{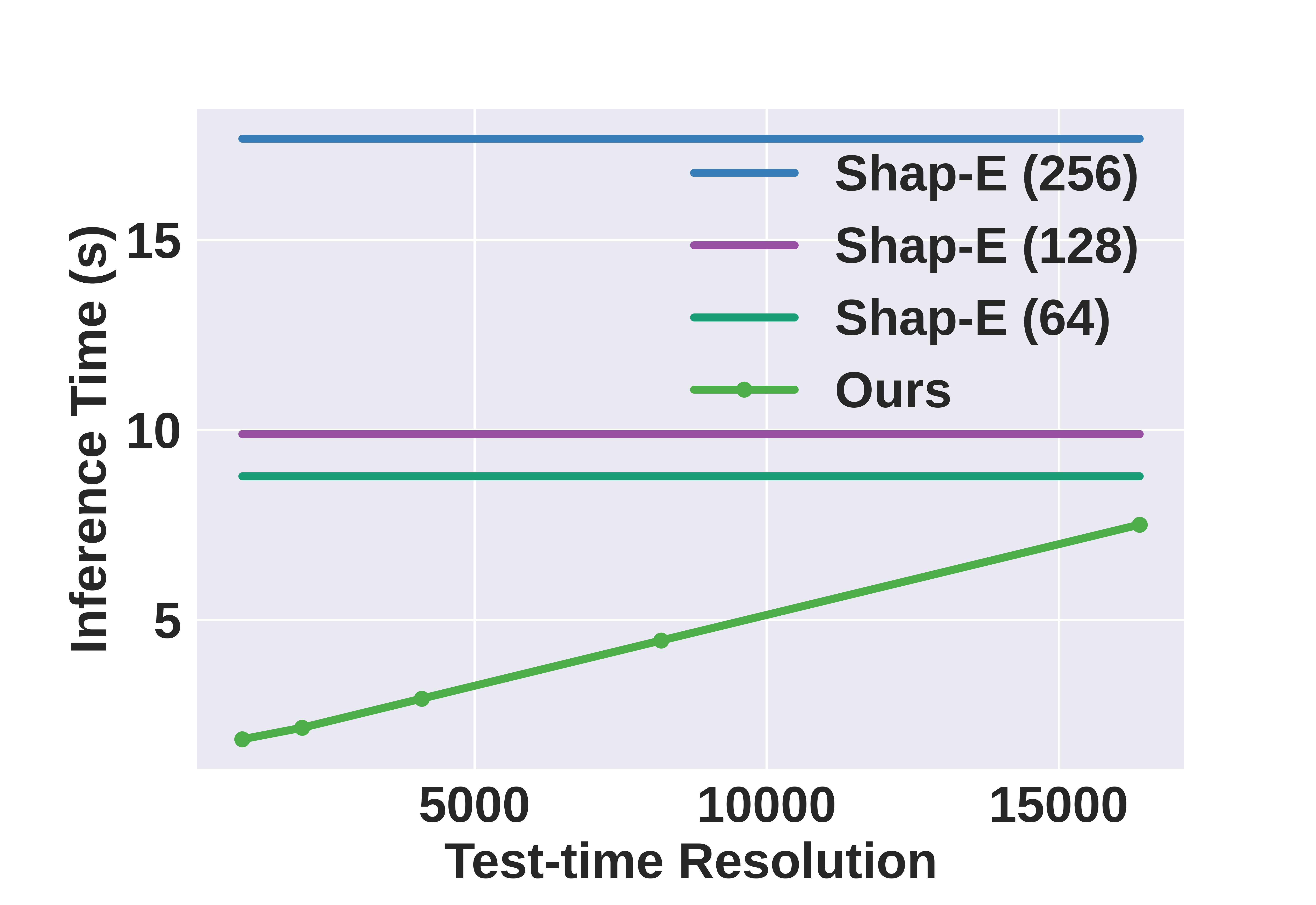}
\includegraphics[width=0.495\linewidth]{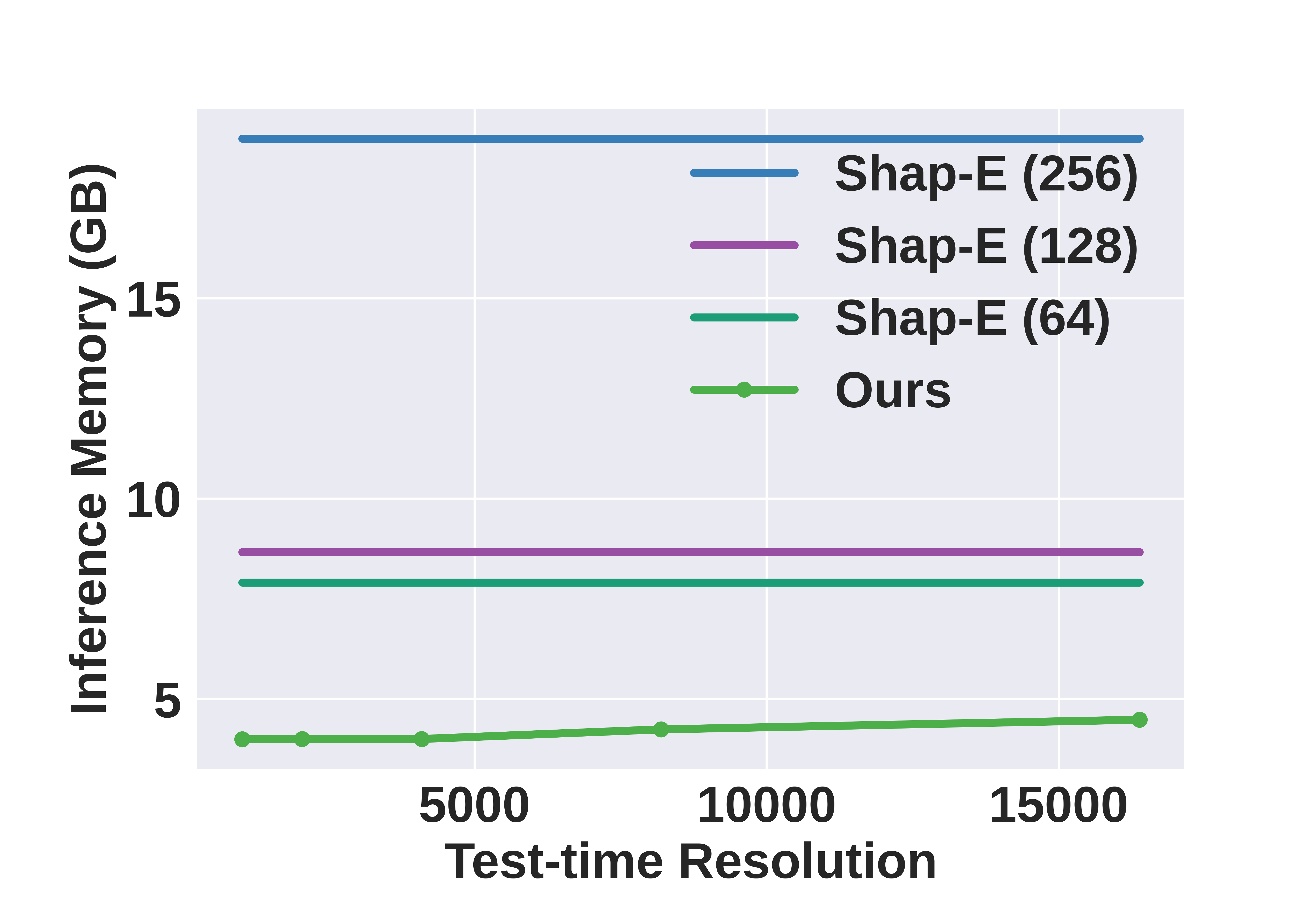}
\caption{\textbf{\method achieves favorable computational complexity even compared with implicit methods such as Shap-E~\cite{jun2023shap}.} The figures show \method is faster and more memory-efficient than Shap-E under a high test-time resolution of 16k.}
\label{fig:shap-e-compare}
\vspace{-2mm}
\end{figure}

We further compare the computational efficiency of PointInfinity to diffusion models with implicit representations. We consider the state-of-the-art implicit model, Shap-E~\cite{jun2023shap}. For a comprehensive comparison, we run Shap-E under different commonly used marching cubes resolutions and show results in~\cref{fig:shap-e-compare}. Our results show that PointInfinity is faster and more memory-efficient than Shap-E.

Overall, \method demonstrates significant advantage in computational efficiency.

\subsection{Ablation Study}

\paragraph{Training Resolution.}
In~\tabref{ablation:train-reso},
we train our model using different training resolutions and report the performance under a test-time resolution of 16k. We can see that \method is insensitive to training resolutions. We choose 1024 as our training resolution to align with Point-E~\cite{nichol2022point}. 

\paragraph{Number of Latent Tokens.}
We next study the impact of representation size (the number of tokens) used in the `latent stream'.
As shown in~\tabref{ablation:latent}, 256 or higher tends to provide strong results, while smaller values are insufficient to model the underlying shapes accurately.
We choose 256 as our default latent token number for a good balance between performance and computational efficiency.

\paragraph{Comparison to A Na\"ive Mixture Baseline.}
Finally, note that a na\"ive way to increase testing resolution without re-training a model is to perform inference multiple times and combine the results.
We compare \method with the na\"ive mixture baseline (denoted `mixture') in~\tabref{ablation:mixture}.
Interestingly, we observe that the mixture baseline sees a slight improvement with higher resolutions, instead of staying constant.
In a more detailed analysis we found that 
mixing multiple inference results reduces the bias and improves the overall coverage, and thus its CD@1k and FS.
Nonetheless, \method performs significantly better, verifying the non-trivial modeling power gained with our design.
Also note that \method is significantly more efficient, because all points share the same fixed-sized latent representation and are generated in one single inference run.

\subsection{Qualitative Evaluation}
\begin{figure*}[t]
    \centering
        \includegraphics[width=1\linewidth]{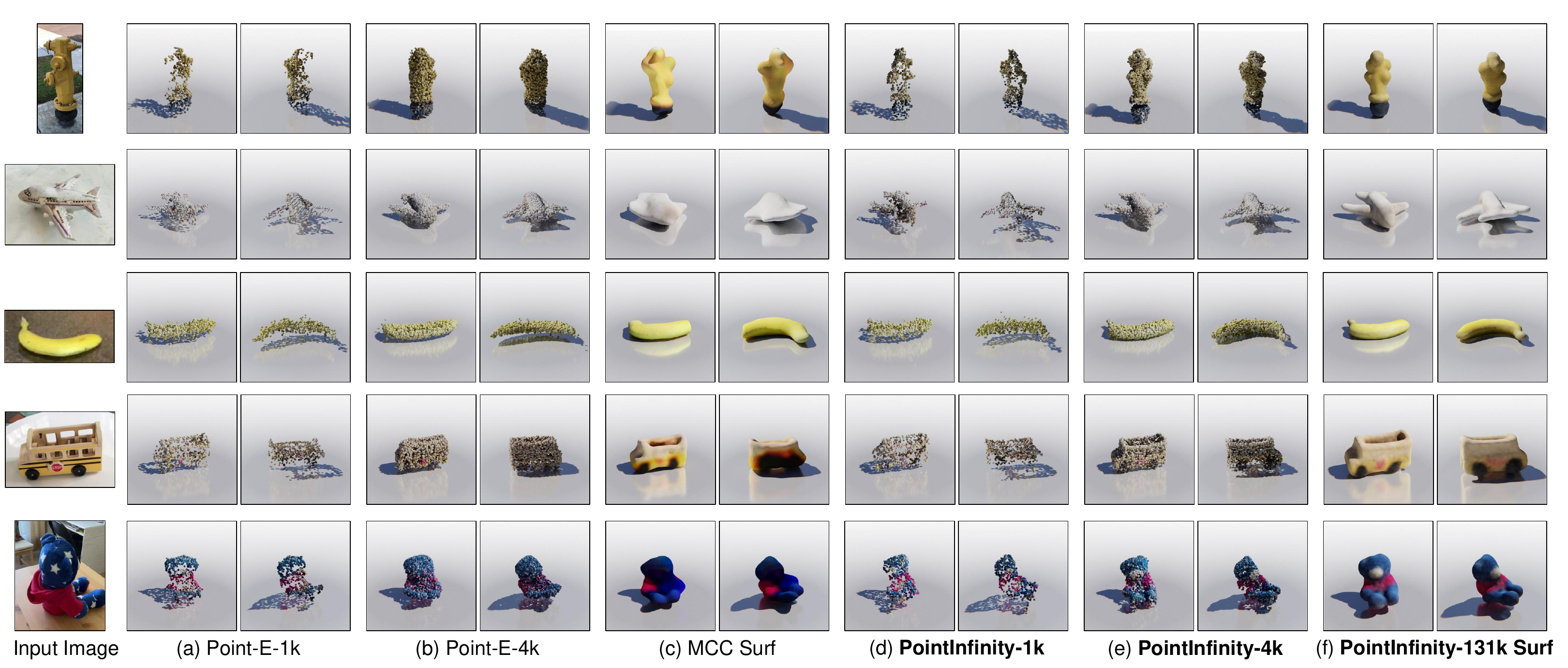}
        \caption{\textbf{Qualitative Evaluation on the CO3D-v2 Dataset~\cite{reizenstein2021common}.} The point clouds generated by our model (column d,e,f) represent denser and more faithful surfaces as resolution increases. On the contrary, Point-E (column a, b) does not capture fine details. In addition, we see that \method obtains more accurate reconstructions from the 131k-resolution point clouds (column f) compared to MCC's surface reconstructions (column c).}
        \label{fig:qualitative-co3d}
        \vspace{-4mm}
\end{figure*}
Here we qualitatively compare \method with other state-of-the-art methods in~\cref{fig:qualitative-co3d}.
Compared to MCC~\cite{wu2023multiview}, we observe that our method generates more accurate shapes and details, 
confirming the advantage of using a diffusion-based point cloud formulation.
Compared to Point-E~\cite{nichol2022point}, \method is able to generate much denser (up to 131k) points, while Point-E generates up to 4k points, which are insufficient to offer a complete shape.
When comparing under the same resolution, we observe that \method enjoys finer details and more accurate shapes than Point-E.
Furthermore, We observe that \method not only achieves high-quality generation results in general, but the generated surface improves as the resolution increases.

\section{Analysis}
\label{sec:analysis}

\begin{table}[t]
    \tablestyle{1.0pt}{1.05}
    \centering
    \begin{tabular}{@{}llx{33}x{33}x{33}x{33}@{}}
    %\hline
    Metric & Method  & 1024 & 2048 & 4096 & 8192 \\ \shline
    \multirow{2}{*}{CD@1k ($\downarrow$)}   
    & Restricted Read & {0.227} & \cellcolor{green!3} 0.225 & \cellcolor{green!5} 0.220 & \cellcolor{green!3} 0.224 \\ %\cline{2-6} 
    & \textbf{Default}    &  {0.227} & \cellcolor{green!40}\textbf{0.197} & \cellcolor{green!55}\textbf{0.186} & \cellcolor{green!70} \textbf{0.181} \\ \hline
    \multirow{2}{*}{CD@full ($\downarrow$)} 
    & Restricted Read & {0.227} & \cellcolor{green!10} 0.211 &  \cellcolor{green!20} 0.196 & \cellcolor{green!30} 0.190 \\ %\cline{2-6}
    & \textbf{Default}    & {0.227} & \cellcolor{green!40} \textbf{0.185} & \cellcolor{green!60} \textbf{0.164} & \cellcolor{green!80} \textbf{0.151} \\ \hline
    \multirow{2}{*}{PSNR ($\uparrow$)} 
    & Restricted Read & {13.37} & \cellcolor{green!3} 13.39 & \cellcolor{green!5} 13.50 & \cellcolor{green!4} 13.49 \\ %\cline{2-6} 
    & \textbf{Default}   & {13.37} & \cellcolor{green!20} \textbf{13.88} & \cellcolor{green!35} \textbf{14.15} & \cellcolor{green!50} \textbf{14.27}
    \end{tabular}
    \caption{\textbf{Analysis of the Resolution Scaling Mechanism.} To verify our hypothesis discussed in \secref{analysis}, we compare our default implementation to a ``Restricted Read" baseline, where the information intake is limited to 1024 tokens, at different test-time resolutions. We see that the performance no longer monotonically improves with resolution, supporting our hypothesis.}
    \label{tab:capping-scaling-perf}
    \vspace{-4mm}
\end{table}

\subsection{Mechanism of Test-time Resolution Scaling}
\label{sec:mechnism}
In \secref{main_results}, we observe that test-time resolution scaling with \method improves the reconstruction quality.
In this section, we provide a set of analysis to provide further insights into this property.

Recall that during diffusion inference, the model input is a linear combination of the Gaussian noise and the output from the previous sampling step.
Our hypothesis is that, increasing the resolution results in a more consistent generation process, because more information are carried out between denoising steps.
With a higher number of input tokens, the denoiser obtains strictly more information on previously denoised results $\bx_t$, and thus $\bx_{t-1}$ will follow the pattern in $\bx_t$ better. 

To verify this hypothesis, we consider a variant of our model, where the read module only reads from a fixed set of $n_\mathrm{train}$ input tokens.
All other $n_\mathrm{test}-n_\mathrm{train}$ tokens' attention weights are set as zero.
The remaining parts of the model are kept unchanged.
As shown in~\tabref{capping-scaling-perf}, after this modification, CD@1k of the model does not improve with resolution anymore. 
Rather, it remains almost constant. 
This result supports that the high information intake indeed leads to performance improvement.

\subsection{Variability Analysis}
\begin{figure}[t]
\centering
	\includegraphics[width=1\linewidth]{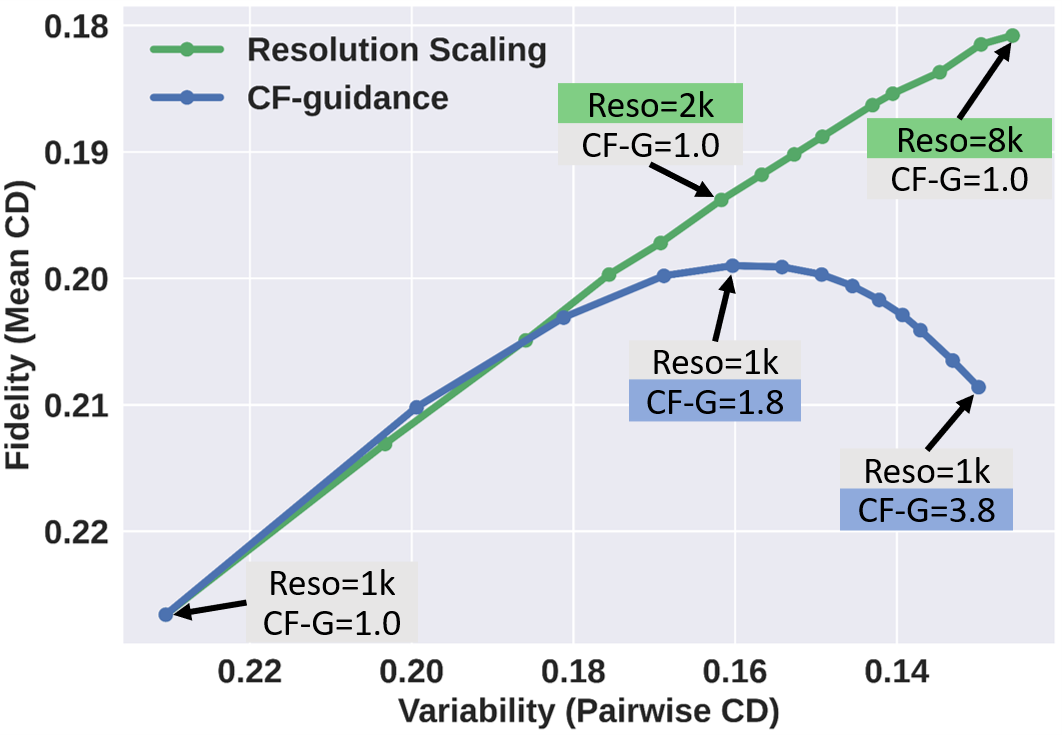}
	\caption{\textbf{Fidelity and Variability Analysis.} We observe that as the resolution increases, the variability of the generated point clouds reduces, due to the stronger condition among the denoising steps. Also note that our test-time resolution scaling achieves a better fidelity-variability trade-off than classifier-free guidance.}
	\label{fig:tradeoff}
    \vspace{-6mm}
\end{figure}

Based on our hypothesis, a potential side effect is a reduced variability, due to the stronger condition among the denoising steps.
To verify this, we evaluate the variability of our sampled point clouds. 
Specifically, for every example in the evaluation set, we randomly generate 3 different point clouds and calculate the average of the pair-wise CD among them, as a measure of the variability. 
In \figref{tradeoff}, we see that when the resolution increases, the variability indeed reduces, supporting our hypothesis.

\subsection{Comparison to Classifier-Free Guidance}
\label{subsec:tradeoff}
The fidelity-variability trade-off observed in resolution scaling is reminiscent of the fidelity-variability trade-off often observed with classifier-free guidance~\cite{ho2022classifier}. 
We compare these two in~\figref{tradeoff}.
As we can see, when the guidance scale is small, classifier-free guidance indeed improves the fidelity at the cost of variability.
However, when the guidance scale gets large, further increasing the guidance hurts the fidelity. 
On the contrary, our resolution scaling consistently improves the sample fidelity, even at very high resolution. 
Moreover, the trade-off achieved by \method is always superior to the trade-off of classifier-free guidance.
\section{Conclusions}
\label{sec:conclusion}
We present \method, a resolution-invariant point diffusion model that efficiently generates high-resolution point clouds (up to 131k points) with state-of-the-art quality.
This is achieved by a two-stream design, where we decouple the latent representation for modeling the underlying shape and the point cloud representation that is variable in size.
Interestingly, we observe that the surface quality in fact \emph{improves} as the resolution increases.
We thoroughly analyze this phenomenon and provide insights into the underlying mechanism.
We hope our method and results are useful for future research towards scalable 3D point cloud generation.
{
    \small
    \bibliographystyle{ieeenat_fullname}
    \bibliography{main}
}


\end{document}